# A New Type of Neurons for Machine Learning

Fenglei Fan, Wenxiang Cong, Ge Wang
Biomedical Imaging Center, BME/CBIS
Rensselaer Polytechnic Institute, Troy, New York, USA

*Abstract* — In machine learning, the use of an artificial neural network is the mainstream approach. Such a network consists of layers of neurons. These neurons are of the same type characterized by the two features: (1) an inner product of an input vector and a matching weighting vector of trainable parameters and (2) a nonlinear excitation function. Here we investigate the possibility of replacing the inner product with a quadratic function of the input vector, thereby upgrading the 1$^{st}$ order neuron to the 2$^{nd}$ order neuron, empowering individual neurons, and facilitating the optimization of neural networks. Also, numerical examples are provided to illustrate the feasibility and merits of the 2$^{nd}$ order neurons. Finally, further topics are discussed.

## I. Introduction

IN the field of machine learning, artificial neural networks (ANNs) especially deep neural networks (CNNs) have recently achieved remarkable successes in various types of applications such as classification, unsupervised learning, prediction, image processing and analysis [1-3]. Excited by the tremendous potential of machine learning, major efforts are being made to improve machine learning methods [4]. An important aspect of the methodological research is how to optimize the topology of a neural network. For example, generative adversarial networks (GANs) were proposed to utilize some deep results of game theory. In particular, the Wasserstein distance was introduced for effective, efficient and stable performance of GAN [5, 6].

To our best knowledge, all ANNs/CNNs are currently constructed with neurons of the same type characterized by the two features: (1) an inner product of an input vector and a matching weighting vector of trainable parameters and (2) a nonlinear excitation function [7-9]. Although these neurons can be interconnected to approximate any general function, the topology of the network is not unique. On another side of this non-uniqueness, *our hypothesis is that the type of neurons suitable for general machine learning is not unique either, which is a new dimension of machine learning research*.

The above hypothesis is well motivated. It is commonly known that the current structure of artificial neurons was inspired by the inner working of biological neurons. Biologically speaking, a core principle is that diversity brings synergy and prosperity at all levels. In biology, multiple types of cells are available to make various organisms [10]. Indeed, many genomic components contribute complementary strengths and compensate for others' weaknesses [11]. Most of cell types are highly dedicated to desirable forms and functions. Synthetic biological research may enhance the biological diversity further. *This observation makes us wonder if we could have multiple types of neurons for machine learning either in general or for specific tasks.*

As the first step along this direction, here we investigate the possibility of replacing the inner product with a quadratic function of the input vector, thereby upgrading the 1$^{st}$ order neuron to the 2$^{nd}$ order neuron, empowering individual neurons, and facilitating the optimization of neural networks.

The model of the current single neurons, which are also referred to as perceptrons, has been applied to solve linearly separable problems. For linearly inseparable tasks, multi-layers of neurons are needed to perform multi-scale nonlinear analysis. In other words, existing neurons can only perform linear classification individually, and the linearly-limited cellular function can be only enhanced via cellular interconnection into an artificial organism. *Our curiosity is to produce such an artificial organism with neurons capable of performing quadratic classification*.

In the next section, we describe the 2nd order neurons. Given $n$ inputs, the new type of neurons could have $3n$ parameters to approximate a general quadratic function, or have $n(n+1)/2$ degrees of freedom to admit any quadratic function. Furthermore, we formulate how to train the 2$^{nd}$ order neuron. In the third section, we present numerical simulation results for fuzzy logic operations. In the last section, we discuss relevant issues and conclude the paper.

## II. Second Order model and Optimization

The structure of a current neuron is shown in Fig. 1, where we define $w_0 := b$ and $x_0 = 1$. The linear function of the input vector produce the output [8]:

$$f(x) = \sum_{i=1}^{n} w_i x_i + b \qquad (1)$$

and then $f(x)$ will be nonlinearly processed, such as by a sigmoid function. Clearly, the single neuron can separate two sets of inputs that are linearly separable. In contrast, for linearly inseparable groups of inputs, the single neuron is subject to classification errors. For example, a single neuron is incapable of simulating the function of the XOR gate.

We introduce a new type of neurons in Fig. 2 where the input vector is turned into two inner products and one norm term for summation before feeding to the nonlinear excitation function.

This work is partially supported by the Clark & Crossan Endowment Fund at Rensselaer Polytechnic Institute, Troy, New York, USA.

Again, for compact notation, we define that $w_{0r} := b_1$, $w_{0g} := b_2$ and $x_0 = 1$. Then, the output function is expressed as:

$$f(x) = (\sum_{i=1}^{n} w_{ir} x_i + b_1)(\sum_{i=1}^{n} w_{ig} x_i + b_2) + \sum_{i=1}^{n} w_{ib} x_i^2 + c. \quad (2)$$

In this embodiment, the threshold is chosen by a sigmoid function. Eq. (2) is quadric, and has the current neuron type as a special case. Due to its added nonlinearity, our proposed neuron is intrinsically superior to the current neuron in terms of representation power such as for classification.

The training algorithm can be formulated for the proposed 2nd order neuron as follows, assuming the sigmoid function [8]

$$\sigma(x) = \frac{1}{1 + \exp(-\beta x)}, \quad (3)$$

where $\beta = 1$ in this pilot study. Let us denote a training dataset of $m$ samples by $X^k = (x_1^k, x_2^k, \cdots, x_n^k)$, along with the ideal output data $y^k$, $k = 1, 2, \cdots, m$. Then, the output of the 2nd order neuron can be modeled as

$$h(X^k, w_r, w_g, w_b, b_1, b_2) = \sigma(f(x))$$
$$= \sigma\left(\left(\sum_{i=1}^{n} w_{ir} x_i^k + b_1\right)\left(\sum_{i=1}^{n} w_{ig} x_i^k + b_2\right) + \sum_{i=1}^{n} w_{ib} (x_i^k)^2 + c\right). \quad (4)$$

Let us define the error function as

$$E(\vec{w_r}, \vec{w_g}, \vec{w_b}, b_1, b_2) = \frac{1}{2} \sum_{k=1}^{m} \left(h(X^k, w_r, w_g, w_b, b_1, b_2) - y^k\right)^2 \quad (5)$$

The error function depends on the structural parameters:
$b_1$, $b_2$, $c$, $\vec{w_r} = (w_{1r}, w_{2r}, \cdots, w_{nr})$, $\vec{w_g} = (w_{1g}, w_{2g}, \cdots, w_{ng})$ and $\vec{w_b} = (w_{1b}, w_{2b}, \cdots, w_{nb})$,

The goal of machine learning is to find optimal parameters to minimize the objective function [9]. The optimal parameters can be found using the gradient descent method with an appropriate initial guess. Thus, we can iteratively update $\vec{w_r}$, $\vec{w_g}$, $\vec{w_b}$, $b_1$, $b_2$ and $c$ in the form of $\alpha = \alpha - \eta \cdot \frac{\partial E}{\partial \alpha}$, where $\alpha$ denotes a generic variable of the objective function, and the step size $\eta$ is typically set between zero and one. The gradient of the object function for any sample can be computed as follows:

$$\frac{\partial E}{\partial w_{ir}} = (h(\vec{x_i}) - y_i) \frac{\partial \sigma}{\partial x} x_i (\sum_{i=1}^{n} w_{ig} x_i + b_2)$$

$$\frac{\partial E}{\partial w_{ig}} = (h(\vec{x_i}) - y_i) \frac{\partial \sigma}{\partial x} x_i (\sum_{i=1}^{n} w_{ir} x_i + b_1)$$

$$\frac{\partial E}{\partial w_{ib}} = 2(h(\vec{x_i}) - y_i) \frac{\partial \sigma}{\partial x} w_{ib} x_i$$

$$\frac{\partial E}{\partial b_1} = (h(\vec{x_i}) - y_i) \frac{\partial \sigma}{\partial x} (\sum_{i=1}^{n} w_{ig} x_i + b_2)$$

$$\frac{\partial E}{\partial b_2} = (h(\vec{x_i}) - y_i) \frac{\partial \sigma}{\partial x} (\sum_{i=1}^{n} w_{ir} x_i + b_1)$$

$$\frac{\partial E}{\partial c} = (h(\vec{x_i}) - y_i) \frac{\partial \sigma}{\partial x}$$

A general optimization flowchart is in Fig. 3.

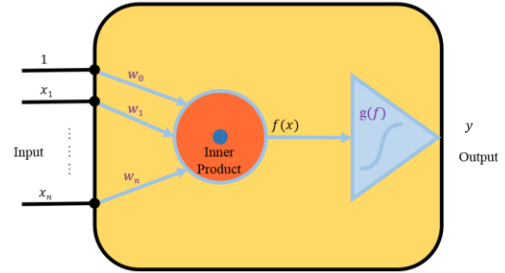

Fig. 1. Current structure of an artificial neuron consisting of a linear inner product and a nonlinear excitation function.

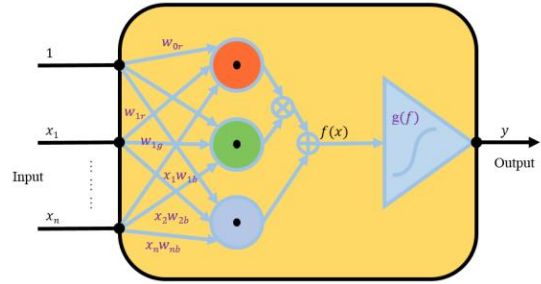

Fig. 2. Exemplary structure of our proposed artificial neuron consisting of a quadratic form and a nonlinear excitation function.

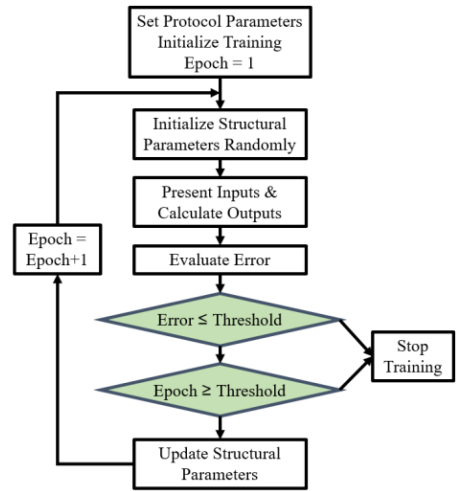

Fig. 3. Flowchart for training the 2nd order neuron in terms of its structural parameters (weights and offsets). Note that we plan to improve this iterative process for higher efficiency.

## III. NUMERICAL RESULTS

In our pilot study, the proposed 2nd order neuron of two input variables was individually applied in fuzzy logic experiments. Extending classic Boolean logic, fuzzy logic was extensively studied and applied over the past decades, which manipulates vague or imprecise logic statements. Compared with Boolean logic, the value of a fuzzy logic variable is on the interval [0, 1], since a practical judgement may not be purely black (false or 0) or white (true or 1). The fuzziness of a logic variable can be easily reflected in a continuous variable by its closeness to either 0 or 1.

For visualization, the color map "cool" in MATLAB was used to represent the functional value at every point. We used "o" for 0 and "+" for 1. With the proposed $2^{nd}$ order neuron, the training process kept refining a quadratic contour to separate labeled points for the highest classification accuracy. By the nature of the $2^{nd}$ order neuron, the contour can be two lines or curves including parabolic and elliptical boundaries.

*A. XOR Gate*

First, the training dataset was simply the XOR logic table, which is not separable by a single neuron of the conventional type. To implement the XOR-gate operation with our proposed $2^{nd}$ order neuron, the initial parameters could be randomly selected in the framework of evolutionary computation. For example, an initial seed were set to $w_r$ =[-0.4, -0.4], $w_g$ =[0.2, 1], $w_b$ =[0, 0], $b_1$ =-0.9095, $b_2$ =-0.6426, $c$=0. The trained logic map is shown in Fig. 4, yielding a perfect result.

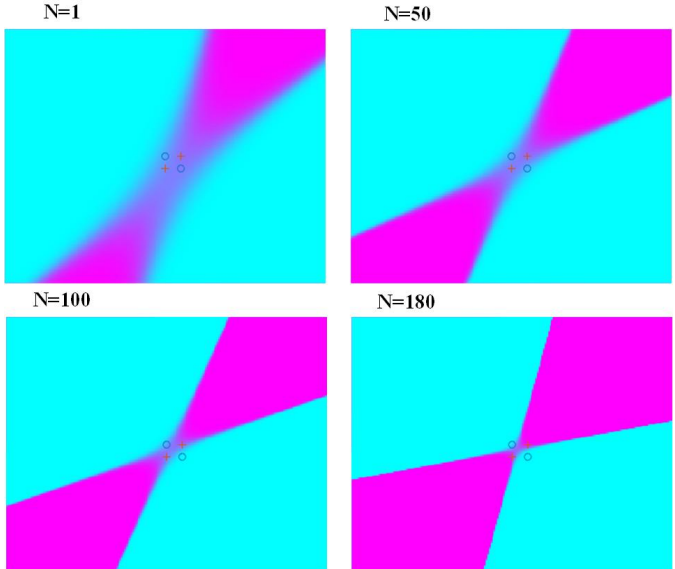

Fig. 4. XOR logic implemented by the proposed $2^{nd}$ order neuron after 180 iterations. After the training, the outputs at [0,0], [0,1], [1,0] and [1,1] are 0.4509, 0.5595, 0.5346 and 0.3111, meaning 0, 1, 1 and 0 respectively relative to the threshold 0.5.

Then, we generated an XOR-like pattern, the initial parameters were set to $w_r$ =[0.07994,-0.2119], $w_g$ =[0.06049,-0.144], $w_b$ =[0 0], $b_1$ =-0.9095, $b_2$ =-0.6426. $c$=0. As shown in Fig. 5, the deformed XOR pattern was perfectly segmented by our proposed $2^{nd}$ order neuron after training.

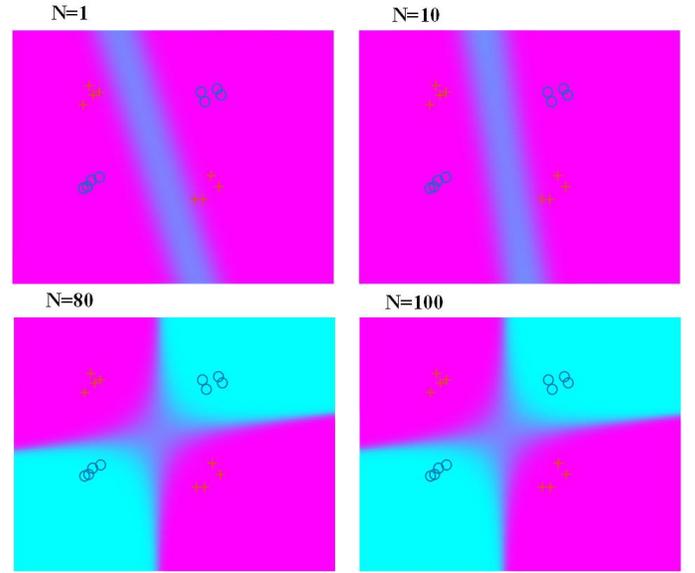

Fig. 5. XOR-like function computed by the proposed $2^{nd}$ order neuron after 100 iterations.

*B. NAND and NOR Gates*

While the XOR gate is exemplary, it is not a universal logic gate. To further show the power of the proposed $2^{nd}$ order neuron, additional datasets were generated that respectively demand fuzzy NAND and NOR operations for correct classification. Then, training steps similar to those in Subsection A were repeated to simulate NAND and NOR operations respectively. In this pilot study, the initial parameters were set to $w_r$ =[0.4, -0.1], $w_g$ =[0.3, 0.1], $w_b$ =[0, 0], $b_1$ =0, $b_2$ =0, $c$=1.3 and $w_r$ =[-1,1], $w_g$ =[1,-2], $w_b$ =[0,0], $b_1$ =-0.5, $b_2$ =1, $c$=0, for NAND and NOR tasks respectively. The results are shown in Figs. 6 and 7.

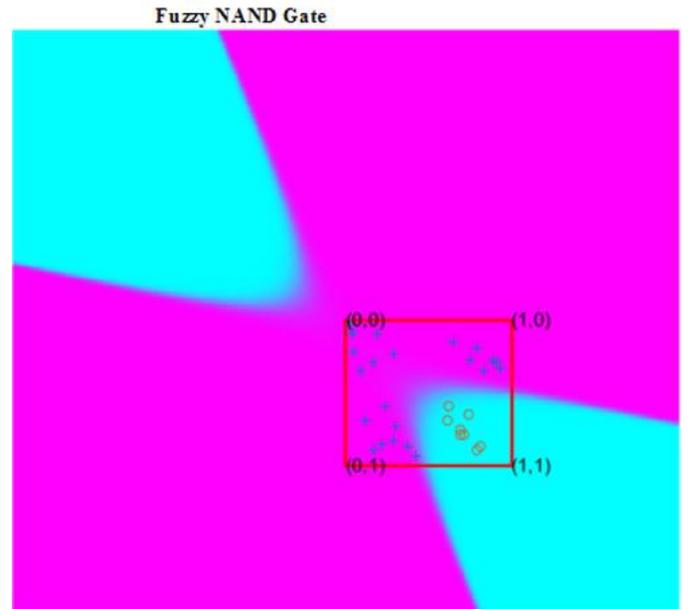

Fig. 6. Fuzzy NAND function implemented by the $2^{nd}$ order neuron after 300 iterations.

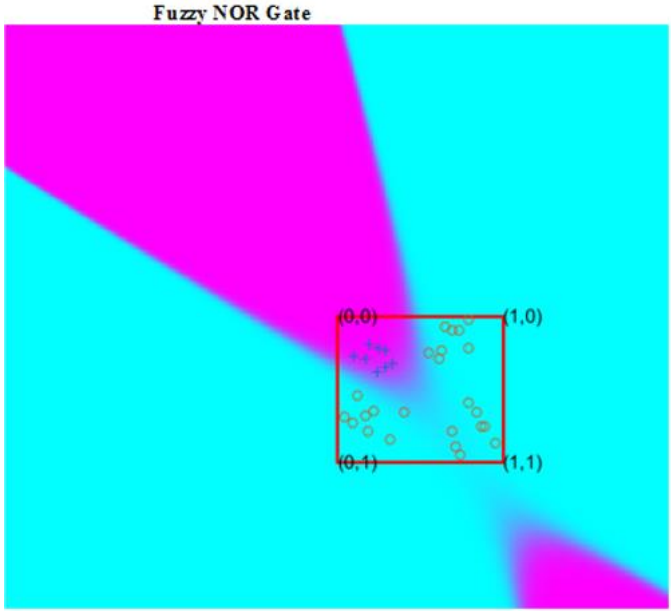

Fig. 7. Fuzzy NOR function of the 2$^{nd}$ order neuron after 100 iterations.

*C. Concentric Rings*

Yet another type of natural patterns linearly inseparable is concentric rings. As an example, we generated two concentric rings, which were respectively assigned to two classes. With the initial parameters $w_r$ =[0.04,0.01], $w_g$ =[0.03,-0.01], $w_b$ =[0,0.4], $b_1$ =0.1, and $b_2$ =0.2, $c$=1.3, our 2$^{nd}$ order neuron was trained, producing an ideal outcome as shown in Fig. 8, which completely separates the inner ring from the outer ring.

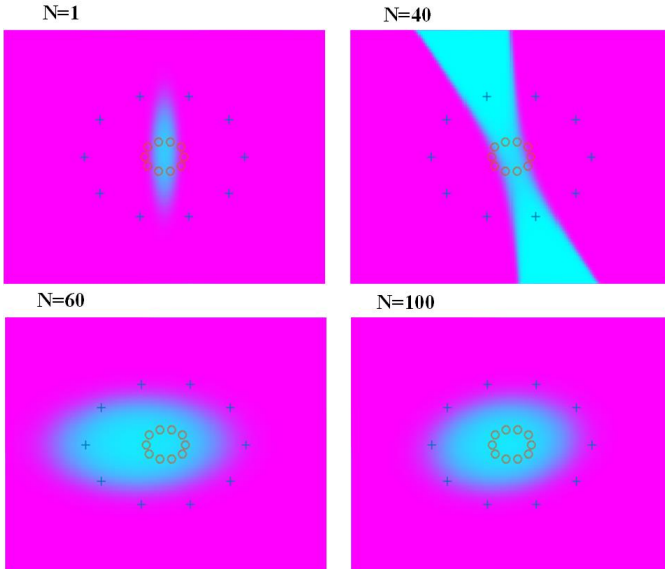

Fig. 8. Perfect classification of concentric rings with the proposed 2$^{nd}$ order neuron.

IV. DISCUSSIONS AND CONCLUSION

As demonstrated in this paper, the proposed 2$^{nd}$ order neuron works well in solving basic fuzzy logic problems, and qualified as a different building block for machine learning. The sufficiency of the proposed neuron type can be trivially argued.

Clearly, the traditional neuron type is a special case of the proposed 2$^{nd}$ order neuron. Since the traditional neuron is sufficient for a general machine learning task, the proposed neuron type should be sufficient as well. Furthermore, the superiority of the proposed neuron type can be argued in analogy to the functional approximation. The traditional neuron linearly synthesizes inputs into a single number, which is the 1$^{st}$ order Taylor approximation. What we have proposed is the 2$^{nd}$ order Taylor approximation. This flexibility should simplify the overall architecture of a neural network in certain types of applications including but not limited to fuzzy logic tasks.

From the perspective of digital/fuzzy logic, the proposed 2$^{nd}$ order neurons can directly implement all typical logic gates with or without fuzzyness, especially the so-called universal gates such as NAND and NOR. This comment is a hint that the 2$^{nd}$ order neurons may be appropriate modules for fuzzy logic processing or soft-computing when logic values are not binary, which means the correct logic judgements cannot be directly made with the common neuron that is the 1$^{st}$ order but can be efficiently made with the proposed 2$^{nd}$ order neuron.

While the proposal 2$^{nd}$ order neuron has important new representation capabilities, the general 2$^{nd}$ order neuron should use more parameters:

$$f(\vec{x}) = \sum_{i,j=1, i\geq j}^{n} a_{ij} x_i x_j + \sum_{k=1}^{n} b_k x_k + c$$

From a training set $\{\vec{x}_p\}$ and $\{y_p\}$, the parameters $\{a_{ij}\}$, $\{b_k\}$ and $c$ can be updated using the gradient descent method in the evolutionary computing framework, where the gradient can be computed as follows:

$$\frac{\partial E}{\partial a_{ij}} = (h(\vec{x}_p) - y_p) \frac{\partial \sigma}{\partial x} x_i x_j,$$

$$\frac{\partial E}{\partial b_k} = (h(\vec{x}_p) - y_p) \frac{\partial \sigma}{\partial x} x_k,$$

$$\frac{\partial E}{\partial c} = (h(\vec{x}_p) - y_p) \frac{\partial \sigma}{\partial x}.$$

This general 2$^{nd}$ order neuron can be similarly trained. For example, this general neuron was adapted for an OR-like operation with the initial parameters $a_{11} = a_{22} = a_{21} = 0.1$, $b_1 = b_2 = 1$, $c = 0.1$, yielding the result in Fig. 8.

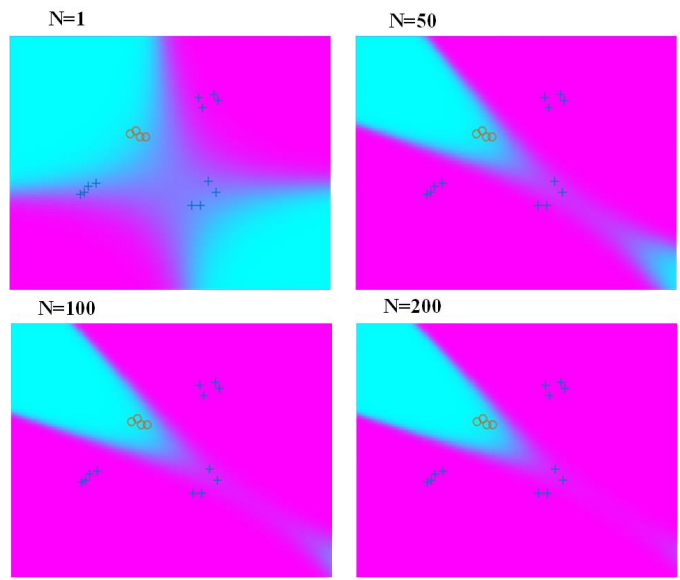

Fig. 8. OR-like pattern classified with the general 2$^{nd}$ order neuron.

Despite the fact that the number of parameters is now in the order of $n^2$, this should not present a major computational challenge when these 2$^{nd}$ order neurons are used in CNNs.

It is underlined that with new types of neurons, we have broadened the space for the network optimization to adjust not only neural interconnections but also neural inner makings. The 2$^{nd}$ order neuron is only an example. Other types of neurons can be imagined as well. Preferred neuron structures may depend on specific applications.

To have a greater impact of the 2$^{nd}$ order neurons, we would need to connect them into multi-layers and compare the power and efficiency of competing networks made with 1$^{st}$ and 2$^{nd}$ order neurons respectively. To train a network consisting of 2$^{nd}$ order neurons, we will formulate a generalized back propagation algorithm and optimize its performance. Importantly, we need to find specific niche applications of 2$^{nd}$ order neurons.

A complete theory remains missing for the neural network consisting of 1$^{st}$ order neurons, and needless to say about theoretical results on the network with 2$^{nd}$ order neurons. However, this challenge is also our opportunities to perform further research on solution optimality, robustness, etc.

In conclusion, a new type of artificial neurons, which we call the 2$^{nd}$ order neurons, has been proposed for an enhanced expressing capability than what the current neuron has. Our pilot results show some encouraging results, but much efforts are needed to demonstrate a real-world utility and establish a rigorous theory.